\documentclass{article}
\usepackage{spconf,amsmath,graphicx}
\usepackage{textcomp}
\usepackage{gensymb}
\usepackage{multirow}
\usepackage{tabulary}
\usepackage{booktabs}

\DeclareMathAlphabet{\pazocal}{OMS}{zplm}{m}{n}
\newcommand\figref{Fig.~\ref}

\newcommand\eref{Equ.~\ref}

\newcommand{\eg}{\textit{e}.\textit{g}.}
\newcommand{\ie}{\textit{i}.\textit{e}.}
\newcommand{\etal}{\textit{et al}.}

\mathchardef\mhyphen="2D 

\title{Towards Pose-invariant Lip-Reading}

\makeatletter
\def\@name{\emph{Shiyang Cheng$^{*,1,3}$,  Pingchuan Ma$^{*,1}$, Georgios Tzimiropoulos$^{2,3}$, Stavros Petridis$^{1,3}$} \\ \emph{Adrian Bulat$^{3}$, Jie Shen$^{1,3}$, Maja Pantic$^{1,3}$}}
\makeatother

\address{$^1$Imperial College London, $^2$University of Nottingham, $^3$Samsung AI Center, Cambridge}

\begin{document}
%
\maketitle
\begin{abstract}
Lip-reading models have been significantly improved recently thanks to powerful deep learning architectures. However, most works focused on frontal or near frontal views of the mouth. As a consequence, lip-reading performance seriously deteriorates in non-frontal mouth views. In this work, we present a framework for training pose-invariant lip-reading models on synthetic data instead of collecting and annotating non-frontal data which is costly and tedious. The proposed model significantly outperforms previous approaches on non-frontal views while retaining the superior performance on frontal and near frontal mouth views. Specifically, we propose to use a 3D Morphable Model (3DMM) to augment LRW, an existing large-scale but mostly frontal dataset, by generating synthetic facial data in arbitrary poses. The newly derived dataset, is used to train a state-of-the-art neural network for lip-reading. We conducted a cross-database experiment for isolated word recognition on the LRS2 dataset, and reported an absolute improvement of 2.55\%. The benefit of the proposed approach becomes clearer in extreme poses where an absolute improvement of up to 20.64\% over the baseline is achieved.
\end{abstract}
\begin{keywords}
lip reading, deep learning, 3DMM
\end{keywords}

\let\thefootnote\relax\footnote{* The first two authors contributed equally.}
\section{Introduction}

Recently, several deep learning approaches for lip-reading \cite{ninomiya2015integration,ngiam2011multimodal,petridis2016deep,sui2014extracting,chung2016lip} have been presented, replacing the traditional feature extraction process by automatically extracting features from the pixels, and significantly outperforming the traditional approaches. The performance has been further improved by the introduction of
end-to-end approaches which attempt to jointly learn the extracted features and perform visual speech classification \cite{petridis2017deepVisualSpeech,chung2016lipSentences,wand2016lipreading,assael2016lipnet}.
%

\looseness -1 
The vast majority of the aforementioned works focused on frontal view lipreading. As a consequence, the performance of such systems degrades in realistic in-the-wild scenarios where the face might not be frontal. To alleviate this, two different approaches have been followed in the literature. The first one trains classifiers using data from all available views in order to build a generic classifier~\cite{lucey2008continuous,chung2017lip}. The second approach applies a mapping to transform features from non-frontal views to the frontal view. Lucey~\etal~\cite{lucey2007Ext} apply a linear mapping to transform profile view features to frontal view features. This approach has been extended to map other views like 30$\degree$, 45$\degree$ and 60$\degree$ to the frontal view~\cite{Estellers2011} or to the 30$\degree$ view~\cite{Lan2012ViewInd}. However, the performance is degraded as the number of features to be generated by the linear mapping increases \cite{lucey2007Ext}.  A similar approach \cite{koumparoulis2018deep} has been presented recently in which the mouth ROIs are frontalised using generative adversarial networks (instead of predicting frontal view features). One recent work~\cite{petridis2017end} on multi-view lip-reading tries to combine multiple views of the mouth to improve the performance. However, as it requires multiple cameras, the usage is limited to certain scenarios like meetings and car environments.

All the above works have been applied to small datasets only. Collecting and annotating a large non-frontal lip-reading database requires tremendous time and efforts. As an alternative, in this work, we present an approach that leverages the 3DMM~\cite{blanz2003face} which, starting from the frontal database of LRW~\cite{chung2016lip}, enables the generation of synthetic lip-reading data in arbitrary poses. This allows the model to be trained on a large range of poses which results in significant performance improvement on non-frontal views. 
%
\begin{figure*}[ht!]
    \centering
    \includegraphics[width=0.9\linewidth]{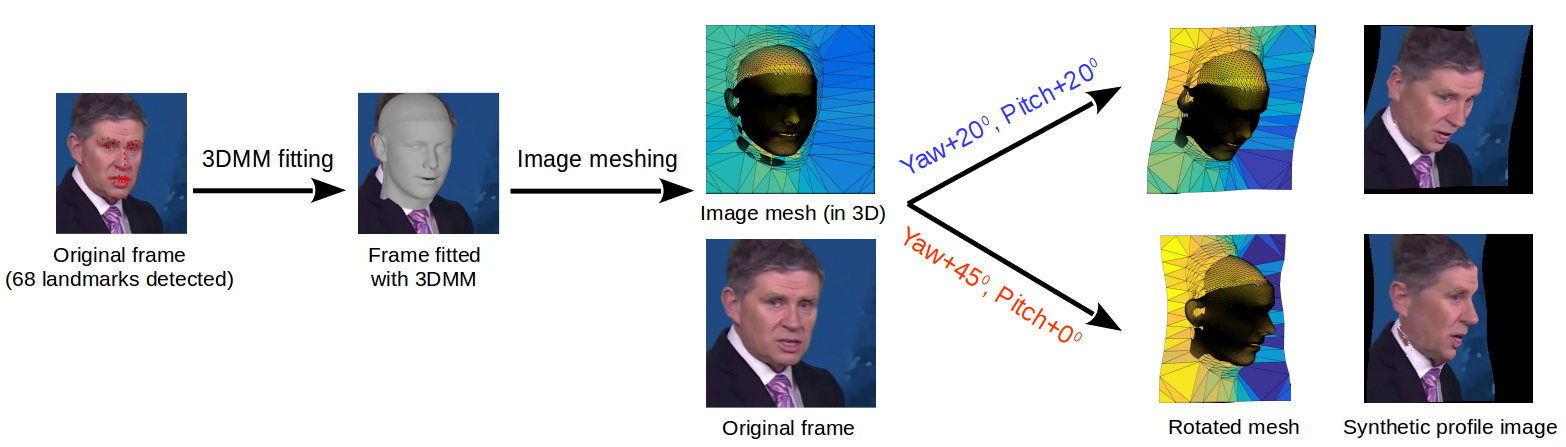}
    \vspace{-13pt}
    \caption{Pose augmentation pipeline.}
    \label{fig:pose_aug_pipeline}
    \vspace{-10pt}
\end{figure*}
The main goal of pose-invariant lip-reading is to reduce the impact of different poses as it is known that the performance decreases when a classifier is trained and tested on different poses.

Our contributions can be summarised as follows: (i) We describe a method to construct a large-pose synthetic database for lip-reading. Our method capitalizes on robust 3DMM fitting~\cite{zhu2019face}, which allows us to take as input a frontal facial image and render it in any arbitrary pose. Using this method, we derive a database that extends the large-scale but mostly frontal database of LRW, which we call LRW in Large Poses (\textbf{LP}). (ii) We investigate the effect of image augmentation as a way to further boost performance. (iii) We use the synthetic database to train a state-of-the-art model \cite{stafylakis2017combining,petridis2018end}, and show that the new model significantly outperforms its counterpart trained on LRW. We conducted a cross-database experiment for isolated word recognition on the LRS2 database and achieved an improvement of 2.55\%. We also show the benefit of the proposed approach in extreme poses, where an improvement of up to 20.64\% is achieved.

\section{Methodology}


\subsection{Pose augmentation}
Our core idea is to generate large-pose lip-reading data by augmenting LRW~\cite{chung2016lip} which is a large-scale but mostly frontal database. A simple but effective approach to generate profile faces is through the use of 3DMM~\cite{blanz2003face}. The pose augmentation pipeline, demonstrated in~\figref{fig:pose_aug_pipeline}, consists of 2 steps: (1) fit the 3DMM into the 2D image; (2) rotate the fitted 3D face to a new angle and render a new image.

\noindent \textbf{3DMM fitting.} Following prior works, we use a combined 3DMM consisting of the Basel~\cite{paysan20093d} and FaceWarehouse~\cite{cao2014facewarehouse} models. A typical 3DMM can be expressed as:
\begin{equation}
    \mathbf{S}_{3D} = \Bar{\mathbf{S}}_{3D} + \mathbf{U}_{id}\mathbf{p}_{id} + \mathbf{U}_{exp}\mathbf{p}_{exp},
    \label{eq:3dmm}
\end{equation}
where $\mathbf{S}_{3D}$ is the reconstructed 3D face, and $\Bar{\mathbf{S}}_{3D}$ is the mean 3D face, $\mathbf{U}_{id}$ and $\mathbf{U}_{exp}$ are the eigenbasis for facial identity and expression respectively, and $\mathbf{p}_{id}$ and $\mathbf{p}_{exp}$ are the corresponding parameters. By applying a weak perspective projection on the 3D model, we project the mesh onto the image:
\begin{equation}
    \mathbf{S}_{2D} = f * \mathbf{C} * \mathbf{R} * \mathbf{S}_{3D} + \mathbf{t}_{2D},
    \label{eq:3dmm_projection}
\end{equation}
where $\mathbf{S}_{2D}$ denotes the 2D coordinates of projected 3D mesh on the image plane, $\mathbf{C}$ is the orthographic projection matrix, $f$, $\mathbf{R}$ and $\mathbf{t}_{2D}$ denote the scale, rotation and translation respectively. We can group the parameters from \eref{eq:3dmm} and~\ref{eq:3dmm_projection} into a single set $\mathbf{p} = \{f, \mathbf{R}, \mathbf{t}_{2D}, \mathbf{p}_{id}, \mathbf{p}_{exp}\}$. 3DMM fitting is then defined as the process of recovering $\mathbf{p}$ given the 3DMM and an input 2D facial image. The goal is to recover $\mathbf{p}$ such that the error between the 2D projection and the given image is minimized. 

Fitting 3DMM into 2D images is a difficult optimization problem, an extensive discussion of which falls beyond the scope of this work. In our case, we use the state-of-the-art 3DDFA method proposed in~\cite{zhu2019face, 3ddfa_cleardusk}, which trains a fully convolutional network to regress the parameters $\mathbf{p}$ from an input 2D image in a cascaded manner. In order to provide a better initialization for the 3DMM fitting, we detect 68 facial landmarks for every frame using FAN~\cite{bulat2017far}, which is also used to accurately crop the input image.

\noindent \textbf{Rendering of new poses.} After fitting the 3DMM to a given facial image, we can use the result to render the same face in a new pose. Unlike most previous face augmentation methods~\cite{zhao2017dual,masi2016we,deng2018uv} that only synthesize profile faces without background context, we generate profile faces while trying to preserve the original image context. This is essential for training the deep lip-reading neural network that can work well with real-world images at test time. To this end, we used the face rendering method of~\cite{zhu2019face}. In particular, the fitted 3D face is used for estimating the depth of anchors in the image background (computed using the method from~\cite{zhu2015high}). Then, the whole image is triangulated to create a new 3D mesh using the estimated depth. Finally, we rotate this mesh in 3D space to generate new facial images in arbitrary pose.

\noindent \textbf{LRW in Large Pose (LP).} We apply the techniques described in the previous sub-sections to obtain a new database for lip reading which we call \textbf{LP}. In particular, the 3DMM is fitted into each frame from each video in LRW, after which we estimate the 3D face pose of each frame. For each video, we randomly select two pose increment angles, one in yaw (-45$\degree$ to 45$\degree$) and another in pitch (-30$\degree$ to 30$\degree$) direction. To avoid rendering occluded facial parts with random contents, we enforce the sign of increment angles to be the same as the pose of video's first frame. Finally, we rotate all the frames of this video with the same increment values and render them into a new video. Although we only augment each sequence once (namely doubling the size of LRW), the data for each word still cover a full and continuous range of poses, because each word contains nearly 1,000 examples.
%
%
\subsection{2D image augmentation}
\label{sec:2d_image_aug}
We investigate whether standard image augmentation techniques, widely used in image classification \cite{he2016deep}, are also beneficial to improving the accuracy of the lip-reading models. To our knowledge, the only augmentation techniques used in lip reading so far are random cropping and flipping. In addition to these techniques, we randomly augment the data during training by applying (a) random scaling from $0.8\times$ to $1.2\times$, (b) random image degradation by downsampling the mouth region to 0.4--0.8 of their size and and then upsampling them back to the original size and (c) randomly placing rectangular noise patches of size 0.1--0.4 of the mouth region.  

\section{Visual Speech Recognition}
\begin{figure}[t]
    \centering
    \includegraphics[width=.5\textwidth]{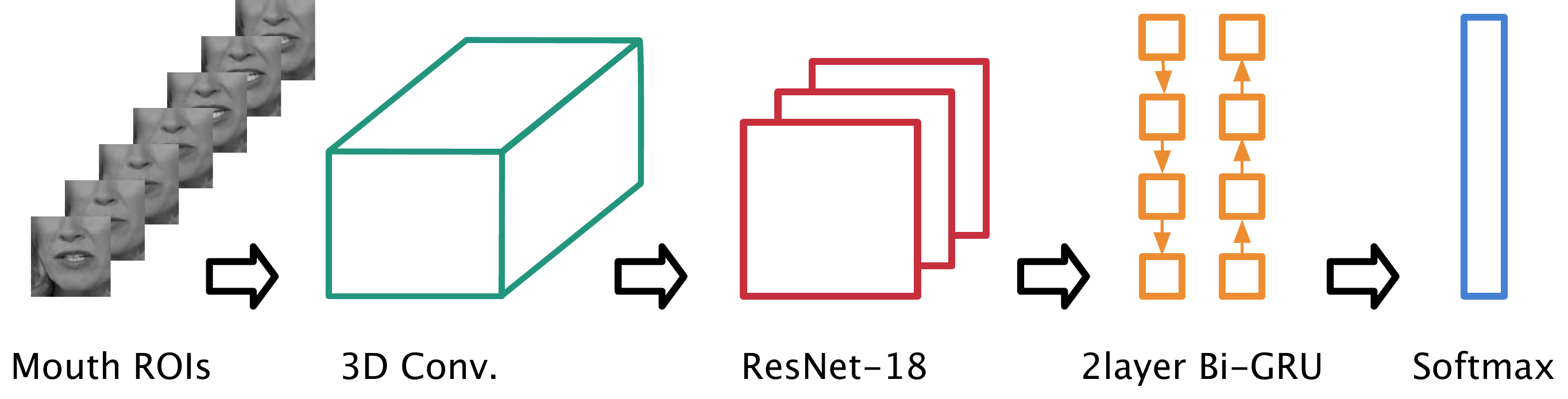}
    \vspace{-10pt}
    \caption{The block diagram of our lip reading model.}
    \label{fig:network}
    \vspace{-10pt}
\end{figure}

The  deep learning model used for visual speech recognition is shown in~\figref{fig:network}. It consists of a residual network (ResNet) \cite{he2016deep} for automatic feature extraction and a 2-layer Bi-GRU to model the temporal dynamics of the features. The architecture is similar to the ones proposed in \cite{stafylakis2017combining,petridis2018end} which achieve state-of-the-art performance on the LRW database. 
The first part of our network performs spatio-temporal convolution, which is capable of capturing the short-term dynamics of the mouth region. It consists
of a convolutional layer with 64 3D kernels of 5 by 7 by 7 size (time/width/height), followed by batch normalization and rectified linear units. This is followed by an 18-layer ResNet to extract the visual features. We have opted for ResNet-18 because preliminary experiments showed it leads to the same performance as ResNet-34 (which is used in the previous studies \cite{stafylakis2017combining,petridis2018end}) but the training time is reduced by 30\%.   Finally, the output of ResNet-18 is fed to a 2-layer BiGRU, each layer consisting of 512 cells.

\section{Experimental Setup}

\subsection{Databases}
\noindent \textbf{Lip Reading in the Wild (LRW) database~\cite{chung2016lip}.} LRW is a large-scale audio-visual database that contains 500 different words from over 1,000 speakers. Each utterance has 29 frames, whose boundary is centered around the target word. The database is divided into training, validation and test sets. The training set contains at least 800 utterances for each class while the validation and test sets contain 50 utterances.

\noindent \textbf{Lip Reading Sentences 2 (LRS2) database~\cite{afouras2018deep}.} LRS2 consists of 224.5 hours of audio-video-text pairs of speaking face collected from BBC TV shows and news. 
It is very challenging due to a large variation in the utterance length and speaker's head pose. 
LRS2 provides word segmentation for the training set, thus we could extract single word utterances from the database to train/test our models.

\subsection{Evaluation protocol}
LRW already contains words so there is no need for further processing. However, for LRS2, we need to extract the individual words first using the word boundaries provided in the training set. 
We find the center frame of each word and select a symmetric 29-frame window around it. In order to filter out some unwanted sequences, we derive some extra rules along this procedure: (1) remove considerably short ($<$ 5 frames) or long segments ($>$ 31 frames); (2) remove segments that appears at the very beginning ($i_{mid} \leq 12$) or the very end ($i_{mid} \geq N_{total} - 12$) of the sentence, where $i_{mid}$ denotes the middle frame of a word and $N_{total}$ is the length of sentence. 

We only select the same 500 words as in LRW, resulting in 60,207 word instances. In the cross-database experiment, we use the whole set from LRS2 as our test data. Additionally, we split this data into training, validation and test sets with a ratio of 8:1:1. Nonetheless, this data is very imbalanced, \eg, some words have over 2,000 examples, while some others have only a few. To balance the split, we limit the number of training and validation examples to 90 and 10 per word, respectively. The final balanced set contains 23175, 3119 and 33909 examples for training, validation and testing respectively, we refer it as $\textbf{LRS2-Ba}$.
\begin{table}[t]
\centering
\small
\begin{tabular}{|c|c|c|c|c|}
\hline
\multirow{2}{*}{Models} & \multicolumn{4}{c|}{Accuracy (\%) on different test sets}\\ \cline{2-5} 
& LRW & LP & LRS2 & LRS2-Ba \\ \hline
M[LRW]     & 82.78 & 69.86 & 57.05 & 54.39 \\ \hline
M[LP]  & 81.67 & 79.08 & 57.25 & 54.43 \\ \hline
M[LRW+LP]    & 83.08 & 79.38 & 58.86 & 56.02 \\ \hline
M[LRW]+Aug2D  & \textbf{83.20} & 72.14 & 58.84 & 56.07 \\ \hline
M[LRW+LP]+Aug2D  & 83.08 & \textbf{79.53} & \textbf{59.60} & 56.78 \\ \hline \hline
M[LRW+LRS2-Ba] & 82.73 & 69.62 & - & \textbf{59.59} \\ \hline
\end{tabular}
\vspace{-5pt}
\caption{Results on the LRW database. \textbf{Aug2D}: 2D image augmentations are applied during training. \textbf{LRS2-Ba}: Balanced partition of LRS2 database.}
\label{tab:result_LRW}
\vspace{-5pt}
\end{table}
\subsection{Data preprocessing}
68 facial points are detected using FAN~\cite{bulat2017far}. All the faces are aligned to a neutral reference frame to remove rotation and scale differences. This is done via an affine transform using 5 landmarks (\ie, eye corners and nose tip). Based on the mouth landmarks, we extract the mouth ROI using a 96$\times$96 bounding box. 
The same procedure is repeated for the entire data to normalise the faces. 

We always include two augmentation techniques that have shown to be useful in lip-reading~\cite{afouras2018deep}, \ie, random cropping (by an 88$\times$88 bounding box) and horizontal flipping (with a probability of 0.5). We optionally include three data augmentation methods introduced in Section~\ref{sec:2d_image_aug} to investigate their impact on visual speech recognition. All these data augmentation methods are applied on video-level, thus the same augmentation setting is configured across all frames.

\subsection{Training details}
Training is divided into 3 phases. We first train a model with a temporal convolutional backend. After this, we replace the backend with a 2-layer Bidirectional Gated Recurrent Unit (Bi-GRU), and we train only the Bi-GRU backend with the weights of the spatiotemporal convolutional layers and ResNet-18 fixed until the model converges. Finally, we train the full model end-to-end. We employ the Adam optimiser \cite{kingma2014adam} with an initial learning rate of 0.0012. Two NVidia Titan 1080Ti with a total batch size of 160 are used.

\section{Results}
\subsection{Overall results}
Results on LRW and LRS2 databases are shown in Table~\ref{tab:result_LRW}. We name our models as follows: (1) \textbf{M[$\cdot$]} denotes the training data composition; (2) A model trained with 2D augmentations (described in Section~\ref{sec:2d_image_aug}) is marked with \textbf{Aug2D}. For instance, \textbf{M[LRW]+Aug2D} is trained using only the original LRW data but with additional image augmentations, \textbf{M[LP]} is trained using only LP data. 
On the other hand, \textbf{M[LRW+LP]} is trained on the combined set of LRW and LP, with the same number of iterations (in one epoch) as that of the baseline model \textbf{M[LRW]}. In this case, we randomly choose examples from either database and we make sure that the total number of training examples remains the same as in the baseline model \textbf{M[LRW]}.

We observe that combining the LP data with the LRW data improves the performance on LRS2 database by 1.81$\%$ (M[LRW+LP]: 58.86\% vs. M[LRW]: 57.05\%). On the other hand, by applying extra 2D image augmentations during training (viz. M[LRW]+Aug2D), we also achieve a better result (58.84\%) than the baseline model. Last, if we employ both pose and 2D image augmentations (viz. M[LRW+LP]+Aug2D), we obtain the best performance in cross-database experiments, \ie, 59.6\% in LRS2 and 56.78\% in LRS2-Ba. Similar conclusions can be reached when testing on LRW and LP databases.

Furthermore, we report one possible \emph{upper-bound} performance (59.59\%) on the balanced LRS2 (\textbf{LRS2-Ba}), which is obtained by a model trained on the combined training set of LRW and LRS2-Ba. We call this model \textbf{M[LRW+LRS2-Ba]}. Results from all other models are also reported for LRS2-Ba test set. Our best model (viz. M[LRW+LP]+Aug2D) results in an absolute improvement of 2.39\% over the baseline model (M[LRW]: 54.39\%) while the upper-bound model (M[LRW+LRS2-Ba]) achieves 59.59\%. Clearly, both pose and 2D image augmentation improve the performance of the model trained on LRW and tested on LRS2-Ba, without any laborious efforts in collecting new data.

\subsection{Pose-wise results on LRS2-Ba}
We further demonstrate pose-wise accuracy achieved by our models on balanced LRS2 data (see 
Tables~\ref{tab:result_yaw} and~\ref{tab:result_pitch}). Specifically, for each word utterance, we estimate the 3D pose of every frame and compute the average pose of the sequence. For simplicity, we take the \emph{absolute value} of poses, based on which, we divide all the sequences into difference pose groups. Taking the yaw angle as an example, we create five groups, \ie, $0\degree \mhyphen 15\degree$, $15\degree \mhyphen 30\degree$, $30\degree \mhyphen 45\degree$, $45\degree \mhyphen 60\degree$ and $60\degree \mhyphen 90\degree$. 

From these results, it can be observed that: (1) Incorporating our synthetic large pose database into the training set improves performance across all pose ranges. This is particularly evident in large poses, for instance, M[LRW+LP] results in an absolute improvement over M[LRW] of 6.43\% and 15.92\% for yaw  in the ranges of $45\degree \mhyphen 60\degree$ and $60\degree \mhyphen 90\degree$, respectively. The absolute improvement in pitch is 20.64\% and 15.79\%, respectively. (2) Applying extra 2D image augmentations (M[LRW]+Aug2D) improves the accuracy most of the time, though the improvement is not as significant as that of M[LRW+LP] in large poses. (3) Surprisingly, in the case of large poses, the models trained on combined LRW and LP data sometimes outperform even the upper-bound model M[LRW+LRS2-Ba]. This showcases the effectiveness and usefulness of our pose augmentation approach. 
\setlength\tabcolsep{1 pt}
\begin{table}[t]
\small
\centering
    \begin{tabular}{|c|c|c|c|c|c|}
    \hline
    \multirow{2}{*}{Models} & \multicolumn{5}{c|}{Accuracy (\%) on different poses}\\ \cline{2-6} 
    & $0\degree \mhyphen 15\degree$ & $15\degree \mhyphen 30\degree$ & $30\degree \mhyphen 45\degree$ & $45\degree \mhyphen 60\degree$ & $60\degree \mhyphen 90\degree$ \\ \hline
    M[LRW]    & 58.11 & 55.18 & 49.62 & 41.73 & 23.07 \\ \hline
    M[LRW]+Aug2D & 60.26 & 55.64 & 50.55 & 44.64 & 29.41 \\ \hline
    M[LRW+LP]    & 59.19 & 55.77 & 50.67 & 48.16 & 38.99 \\ \hline
    M[LRW+LP]+Aug2D  & 59.69 & 56.24 & 52.86 & \textbf{49.54} & \textbf{39.68} \\ \hline \hline
    M[LRW+LRS2-Ba]  & \textbf{63.42} & \textbf{59.24} & \textbf{54.29} & 49.34 & 35.76 \\ \hline 
    \end{tabular}
    \vspace{-5pt}
    \caption{LRS2-Ba test accuracy (\%) divided by Yaw angle.}
    \label{tab:result_yaw}
    \vspace{-5pt}
\end{table}
\begin{table}[t]
\small
\centering
    \begin{tabular}{|c|c|c|c|c|c|}
    \hline
    \multirow{2}{*}{Models} & \multicolumn{5}{c|}{Accuracy (\%) on different poses}\\ \cline{2-6} 
    & $0\degree \mhyphen 15\degree$ & $15\degree \mhyphen 30\degree$ & $30\degree \mhyphen 45\degree$ & $45\degree \mhyphen 60\degree$ & $60\degree \mhyphen 90\degree$ \\ \hline
    M[LRW] & 55.77 & 44.86 & 28.22 & 9.52 & 7.89 \\ \hline
    M[LRW]+Aug2D & 57.35 & 47.2 & 32.78 & 9.52 & 10.53 \\ \hline
    M[LRW+LP] & 57.08 & 48.28 & \textbf{38.59} & \textbf{30.16} & \textbf{23.68} \\ \hline
    M[LRW+LP]+Aug2D & 57.77 & 49.95 & 37.34 & 26.98 & 21.05 \\ \hline \hline
    M[LRW+LRS2-Ba] & \textbf{60.79} & \textbf{51.6} & 34.44 & 22.22 & 10.53 \\ \hline 
    \end{tabular}
    \vspace{-5pt}
    \caption{LRS2-Ba test accuracy (\%) divided by Pitch angle.}
    \label{tab:result_pitch}
    \vspace{-5pt}
\end{table}
\section{Conclusion}
We have presented a method for pose-invariant lip-reading by constructing large-pose synthetic data. The proposed approach is based on 3DMM which allows us to take a frontal facial image and render the face in any arbitrary pose. Augmenting the training set with this method results in improved performance when training on the mostly frontal LRW database and testing on the LRS2 database which contains a variety of poses. It is worth pointing out that a substantial improvement is observed in extreme poses, beyond 45$\degree$ in yaw and pitch. In future work we will investigate the performance of the proposed approach on other databases with more extreme poses like LRS3 and on continuous visual speech recognition.

\bibliographystyle{IEEEbib}
\bibliography{egbib}

\end{document}